\pdfoutput=1

\documentclass[11pt]{article}
\usepackage[dvipsnames]{xcolor}

\usepackage[preprint]{acl}

\usepackage{times}
\usepackage{latexsym}

\usepackage[T1]{fontenc}

\usepackage[utf8]{inputenc}

\usepackage{microtype}

\usepackage{inconsolata}

\usepackage{graphicx}
\usepackage{tabularx}
\usepackage{multirow}
\usepackage{amsfonts}
\usepackage{amssymb}
\usepackage{xspace}
\usepackage{lipsum}
\usepackage{adjustbox}
\usepackage{booktabs}
\usepackage{makecell}
\usepackage{url}
\usepackage{enumitem}
\setlist[enumerate]{topsep=0pt, partopsep=0pt, parsep=0pt, itemsep=0pt}
\setlist[itemize]{topsep=3pt, partopsep=0pt, parsep=1pt, itemsep=3pt, leftmargin=*}
\usepackage{listings}
\usepackage{xcolor}
\usepackage{inconsolata}
\usepackage[most]{tcolorbox}
\usepackage{caption}
\usepackage{float}
\usepackage{pgfplots}
\pgfplotsset{compat=1.17}
\usepackage{pgfplotstable}
\usepackage{amsmath}
\usepackage{amsthm}
\usetikzlibrary{calc}

\usetikzlibrary{arrows.meta, positioning, shapes.geometric}


\definecolor{tbdred}{RGB}{150,35,35}

\newcommand{\formalizer}{VLM-as-formalizer}
\newcommand{\planner}{VLM-as-planner}
\newcommand{\method}{VLM-as-perceiver}
\newcommand{\gemma}{Gemma-4-31B}
\newcommand{\qwen}{Qwen3.5-122B-A10B}
\newcommand{\gemini}{Gemini-3.1-Pro}

\lstdefinestyle{mystyle}{
    basicstyle=\ttfamily\small,
    breaklines=true,
    frame=single,
    backgroundcolor=\color{gray!5},
    keywordstyle=\color{black},
    commentstyle=\color{gray},
    stringstyle=\color{black},
    showstringspaces=false,
    tabsize=2,
    rulecolor=\color{black},
    framesep=6pt, 
    columns=fullflexible,
}
\title{Isolating Perception for Multi-View Planning under\\ Occlusion-Induced Partial Observability}

\author{
  {\bfseries Kevin Nguyen \enspace Ceyhun Efe Kayan \enspace Feng Liu \enspace Li Zhang} \\[0.2em]
  Drexel University \\
  {\tt kevin.nguyen267@gmail.com | Harry.Zhang@drexel.edu}
}

\begin{document}
\maketitle

\begin{abstract}
Vision-language models (VLMs) are beginning to saturate visual planning benchmarks, yet they fail to resolve depth ambiguity and degrade sharply with problem size under occlusion-induced partial observability (PO). We introduce BW-PO based on the Blocksworld domain, a controlled benchmark that reduces existing VLM planners and formalizers to near-zero performance, and propose \method{}, a paradigm that narrows the VLM's role to local perception while delegating cross-view evidence fusion and plan generation to external solvers. Across model-specific challenging subsets of BW-PO, this raises plan success from 1\% to 97\% for \gemma{}, 0\% to 95\% for \qwen{}, and 3\% to 84\% for \gemini{}. Separating local visual extraction from cross-view evidence fusion exposes two bottlenecks that are otherwise entangled: fusion under depth ambiguity, and completeness of visual extraction under scene density.\footnote{Our code and data can be found at \url{https://github.com/kevinnguyen267/isolating-perception}.}
\end{abstract}

\begin{figure*}[t]
    \centering
    \includegraphics[width=\textwidth]{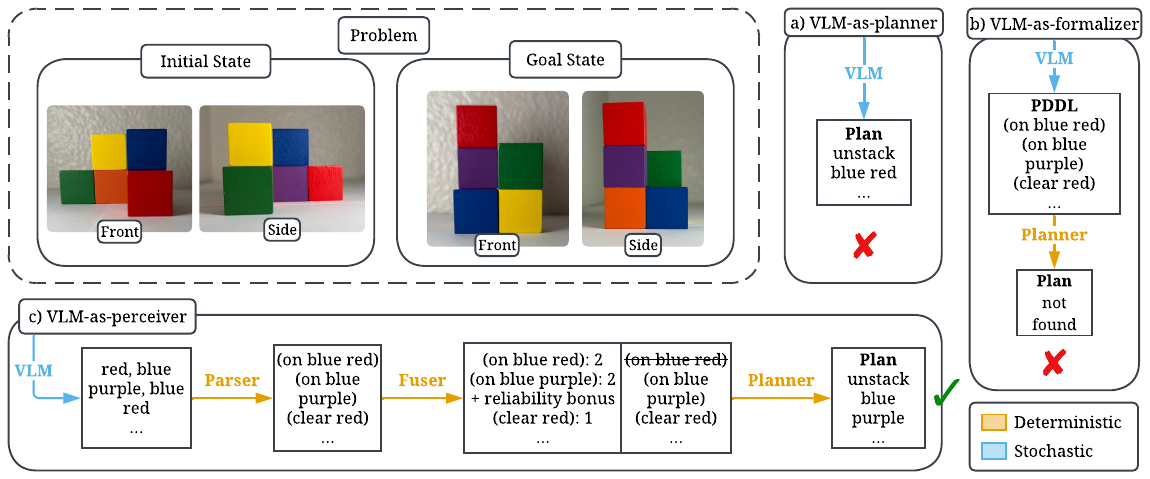}
    \caption{A BW-PO-DEPTH problem where each view cannot individually determine whether the blue block rests on the red block or on the purple block that the red block occludes. In our proposed method (c), the conflicting local claims are fused by anchoring on the purple block that has no conflicting relations. Across (a)--(c), progressively more of the pipeline moves from stochastic to deterministic components.}
    \label{fig:main_figure}
\end{figure*}

\section{Introduction}
\label{sec:introduction}
In visual planning, vision-language models (VLMs) must jointly perform perception, state estimation, and action generation \citep{hu2023look}. This entanglement makes errors difficult to localize and the resulting plans are often plausible but inconsistent with environment dynamics \citep{he2025visionlanguagemodelsplan}. An alternative is to use these models as \emph{formalizers}, translating observations into formal representations, such as the Planning Domain Definition Language (PDDL), which classical planners can then solve deterministically \citep{10611112}.

The central bottleneck in this approach is therefore not action generation, but the accurate extraction of the objects, states, and relations required to formalize a planning problem. This challenge is particularly acute under multi-view partial observability (PO), where each observation provides incomplete evidence about the environment. Recent work has only begun to study planning under such conditions, particularly when partial observability arises from restricted local fields of view and an unexplored environment \citep{nayak2024longhorizon}.

Perception research has followed a complementary path, isolating several challenges to accurate state extraction. These include reasoning about occluded objects \citep{pothiraj2025capture,yun2026integratingmultimodallargelanguage}, resolving the depth ambiguity of which objects lie in front of others \citep{vo2026depthbettertoldshown}, and characterizing how perceptual accuracy degrades as the number of objects increases \citep{sengupta2026visionlanguagemodelscountsynthetic}. Because these studies consider one image at a time, they do not address the additional demands of multi-view state recovery: cross-view correspondences, viewpoint alignment, and the fusion of conflicting evidence. Current models remain weak on all three  \citep{wang2025muirbench,feng2026seeing}.

This gap motivates our study of multi-view planning under occlusion-induced partial observability, where accurate state extraction requires both cross-view integration and robust perceptual reasoning. We introduce \textbf{BW-PO}, built on the widely used Blocksworld domain \citep{ipc1998}, the first benchmark to evaluate this setting along two orthogonal dimensions of difficulty: depth ambiguity and increasing scene complexity as the number of objects grows. Its views are fixed, reflecting real systems whose cameras are fixed or only partially mobile and thus cannot resolve ambiguity through active perception. Both existing approaches, \planner{} and \formalizer{}, degrade to near zero on its most challenging subsets. We therefore propose \textbf{\method{}}. Unlike prior work that separates perception from reasoning, either by using language as the perceptual representation \citep{pan-etal-2024-langnav} or by keeping reasoning and action selection within an LLM \citep{zhou2023navgptexplicitreasoningvisionandlanguage}, it separates local visual extraction from cross-view evidence fusion by narrowing the VLM's role to perception within a single view and delegating fusion to a constraint optimizer and plan generation to a classical planner. This raises performance from \textbf{1\% to 97\%} for \gemma{}, \textbf{0\% to 95\%} for \qwen{}, and \textbf{3\% to 84\%} for \gemini{}, on model-specific challenging subsets. Because fusion becomes deterministic, we can also attribute failure to a specific stage along each axis: fusion under depth ambiguity, and completeness of visual extraction under scene density.

\section{Benchmarks}
\label{sec:benchmarks}

Recent benchmarks such as ProDG \citep{10611112}, BLOCKSWORLD-REAL \citep{he2025visionlanguagemodelsplan}, and ViPlan-BW \citep{merler-etal-2026-viplan} extend the text-based Blocksworld domain, which VLMs have since largely saturated. BW-PO instead comprises six datasets of $110$ problems each, spanning two orthogonal difficulty dimensions: depth ambiguity in BW-PO-DEPTH and increasing object count $N$ in BW-PO-$N$.

Each BW-PO state $s$ is a front- and a side-view image; occlusion leaves each view revealing only part of the scene, so recovering $s$ requires correspondences across viewpoints and fusion of evidence from both images. A planning problem $p=(s_{\mathrm{init}},s_{\mathrm{goal}})$ pairs an initial and a goal state, each given by its own front- and side-view images, and the task is to generate a valid sequence of actions transforming $s_{\mathrm{init}}$ into $s_{\mathrm{goal}}$. Both states are thus subject to partial observability.

BW-PO-DEPTH introduces depth ambiguity within each view: it is not apparent from a single image whether a block $x$ rests directly on a visible block $y$ or on another block occluded by $y$. The alternative viewpoint supplies further evidence, but the two observations induce competing interpretations that must be resolved to recover the true state, as shown in Figure~\ref{fig:main_figure}. This dataset therefore evaluates whether a method can integrate multi-view evidence while reasoning robustly about occlusion and depth.

BW-PO-$N$, where $N\in\{6,12,20,40,100\}$, instead evaluates how performance scales with scene density, well beyond the six blocks of existing Blocksworld visual-planning benchmarks.

\begin{table*}[t]
\centering
\small
\setlength{\tabcolsep}{5pt}
\resizebox{\textwidth}{!}{%
\begin{tabular}{@{}l*{3}{ccc}@{}}
\toprule
\multicolumn{1}{c}{}
& \multicolumn{3}{c}{Gemma-4-31B}
& \multicolumn{3}{c}{Qwen3.5-122B-A10B}
& \multicolumn{3}{c}{Gemini-3.1-Pro} \\
\cmidrule(lr){2-4}
\cmidrule(lr){5-7}
\cmidrule(lr){8-10}

\multicolumn{1}{c}{Dataset}
& Planner & Formalizer & Perceiver
& Planner & Formalizer & Perceiver
& Planner & Formalizer & Perceiver \\
\midrule

BW-PO-DEPTH
& 1\% & 0\% & \textbf{97\%}
& 0\% & 0\% & \textbf{95\%}
& 73\% & 71\% & \textbf{100\%} \\

BW-PO-6
& 82\% & 62\% & \textbf{99\%}
& \textbf{100\%} & \textbf{100\%} & \textbf{100\%}
& 95\% & \textbf{100\%} & \textbf{100\%} \\

BW-PO-12
& 56\% & 20\% & \textbf{86\%}
& 45\% & 50\% & \textbf{51\%}
& 81\% & 95\% & \textbf{100\%} \\

BW-PO-20
& 2\% & 2\% & \textbf{22\%}
& 0\% & 0\% & \textbf{6\%}
& 29\% & 93\% & \textbf{100\%} \\

BW-PO-40
& 0\% & 0\% & \textbf{1\%}
& \textbf{0\%} & \textbf{0\%} & \textbf{0\%}
& 28\% & 89\% & \textbf{90\%} \\

BW-PO-100
& \textbf{0\%} & \textbf{0\%} & \textbf{0\%}
& \textbf{0\%} & \textbf{0\%} & \textbf{0\%}
& 1\% & 3\% & \textbf{84\%} \\

\bottomrule
\end{tabular}%
}
\caption{Plan success rates across all datasets, methods, and models. VLM-as-perceiver achieves or ties for the highest performance across all model--dataset pairs.}
\label{tab:main-results}
\end{table*}

\section{Methods}
\label{sec:methods}

\paragraph{\planner{} (Figure~\ref{fig:main_figure}a).}
All three methods are given the object set. \planner{} additionally receives the available actions and both states at once, each observed from two views, and handles the entire pipeline: it must perceive objects and relations, fuse the views, and search for a valid action sequence. An error at any stage invalidates the plan, and the direct output gives little indication of whether the failure originated in perception, state estimation, or search.

\paragraph{\formalizer{} (Figure~\ref{fig:main_figure}b).}
The VLM is instead given the predicate vocabulary and invoked once per state, translating the two views into a grounded symbolic state to be solved by Fast Downward \citep{Helmert_2006}. Cross-view evidence fusion remains the model's responsibility, so errors committed to the symbolic state before the planner is invoked cannot be corrected downstream.

\paragraph{\method{} (Figure~\ref{fig:main_figure}c).}
The VLM is given neither actions nor predicates and is restricted to local perception within a single view; fusion, formalization, and planning are delegated to the deterministic \emph{Parser}, \emph{Fuser}, and \emph{Planner} stages. It is invoked once per state: for each view $v\in V=\{\mathrm{front},\mathrm{side}\}$ the VLM independently reports a set of vertical stacks $\Sigma_v$, where each $\sigma=(b_1,\dots,b_k)\in\Sigma_v$ lists blocks bottom to top. We write $O$ for the set of blocks reported in any view; unmentioned blocks are absent from the resulting state.

\textbf{Parser.} Each stack is decomposed into $k-1$ relational claims, one per adjacent pair, and two empty-slot claims, for the cells below $b_1$ and above $b_k$. Claims are tallied rather than deduplicated, so a claim made by both views counts twice:
\begin{align*}
    m(x,y)&=\textstyle\sum_{v}\bigl|\{\sigma\in\Sigma_v : y\!\prec\!x \text{ in }\sigma\}\bigr|,\\
    e(x,\mathrm{below})&=\textstyle\sum_{v}\bigl|\{\sigma\in\Sigma_v : b_1=x\}\bigr|,\\
    e(x,\mathrm{above})&=\textstyle\sum_{v}\bigl|\{\sigma\in\Sigma_v : b_k=x\}\bigr|,
\end{align*}
where $y\!\prec\!x$ means $y$ immediately precedes $x$ bottom-to-top.

\textbf{Fuser.} The tallies define a weighted MaxSMT problem solved with Z3 Optimize \citep{10.1007/978-3-540-78800-3_24, 10.1007/978-3-662-46681-0_14}. Each $x\in O$ receives an integer stack index $s(x)$ and height $h(x)$, so that $\mathrm{On}(x,y)\equiv s(x)=s(y)\wedge h(x)=h(y)+1$, subject to the single hard constraint that no cell is occupied twice,
\begin{align*}
    \forall x\neq y:\quad s(x)\neq s(y)\ \vee\ h(x)\neq h(y).
\end{align*}
All claims are soft, with base weights $w(x,y)=2\,m(x,y)$ and $w_{0}(x,d)=e(x,d)$: an observed contact outweighs a conflicting claim that one of its cells is empty, so empty-slot evidence breaks ties without overriding an observed relation.

To resolve competing claims that base weights alone cannot separate, we privilege uncontested local structure. Writing $U(x)=\{y : m(y,x)>0\}$ and $D(x)=\{y : m(x,y)>0\}$ for the candidate blocks observed immediately above and below $x$, and $O^{+}\subseteq O$ for the blocks appearing in at least one relational claim, the reliable blocks are those with an unambiguous neighborhood,
\begin{equation*}
    R=\bigl\{x\in O^{+} \bigm| |U(x)|\leq 1\ \wedge\ |D(x)|\leq 1\bigr\},
\end{equation*}
and each reliable endpoint earns a bonus $B=1+\sum_{(x,y)}w(x,y)$, giving:
\begin{align*}
    \hat{w}(x,y)=w(x,y)+B\,\bigl|\{x,y\}\cap R\bigr|.
\end{align*}
A single bonus outweighs every relational claim combined, so a relation anchored by an undisputed block wins its cell regardless of the opposing evidence, while base weights still discriminate among relations with equal bonuses. Letting $\mathrm{Empty}(x,d)$ denote that the cell adjacent to $x$ in direction $d$ is unoccupied, the solver returns the placement $(s^{\ast},h^{\ast})$ maximizing
\begin{equation*}
\begin{split}
    &\sum_{(x,y)\in C} \hat{w}(x,y)\,\mathbf{1}\bigl[\mathrm{On}(x,y)\bigr]\\
    &\quad{}+\sum_{(x,d)\in E} w_{0}(x,d)\,\mathbf{1}\bigl[\mathrm{Empty}(x,d)\bigr].
\end{split}
\end{equation*}

The gripper is resolved separately by a precedence rule $g$ over per-view votes, detailed in Appendix~\ref{app:gripper}. Because no gripper is visible in any of our scenes, this always yields \texttt{handempty}.

\textbf{Planner.} The placement maps deterministically to the target predicate vocabulary: \texttt{on} for a block with a selected support and \texttt{ontable} otherwise, \texttt{clear} for a block with no selected neighbor above, and \texttt{handempty} or \texttt{handfull} with \texttt{holding} according to $g$. The state is then passed to the same classical planner used by \formalizer{}.

\section{Experiments}
\label{sec:experiments}

\paragraph{Setup.} We evaluate each method with three models: \gemma{} \citep{gemmateam2026gemma4}, \qwen{} \citep{qwen3.5}, and \gemini{}. Our metric is \emph{plan success rate}, the percentage of generated plans that are valid for the corresponding planning problem. Plans are validated with the sequential plan validator of the Unified Planning library \citep{unified_planning_softwarex2025} against the ground-truth domain and problem files. This avoids comparison against a single reference plan, since multiple action sequences may correctly transform the initial state into the goal state, and avoids direct comparison against the ground-truth problem file, since different symbolic formalizations can represent equivalent states and support valid plans. Plan success thus measures symbolic validity; no plans are physically executed.

\paragraph{BW-PO-6 is saturated.} \qwen{} reaches 100\% plan success with all three methods and \gemini{} with both \formalizer{} and \method{} (Table~\ref{tab:main-results}). Six-block scenes therefore discriminate little among current models, even after introducing occlusion-induced partial observability, motivating the harder conditions below.

\paragraph{Depth ambiguity exposes failures in cross-view evidence fusion.} BW-PO-DEPTH produces competing interpretations across views, and only \method{} resolves them reliably, solving 97\% of instances with \gemma{}, 95\% with \qwen{}, and 100\% with \gemini{}. Both existing approaches collapse on the open-weight models, with \planner{} and \formalizer{} reaching 1\% and 0\% with \gemma{} and 0\% with \qwen{}. The gap narrows with \gemini{} (73\% and 71\%) but \method{} stays ahead by nearly 30 points.

To locate the source of these failures, we label each block placement by whether the two views describe it consistently, using the per-view descriptions recovered from a \method{} run (implementation in Appendix~\ref{app:error_analysis}). Across these descriptions, the two views disagree on $31\%$ of placements, and in every case the truth is one of the two competing claims, making each disagreement a binary choice at $50\%$ chance. Because a valid plan requires both states to be recovered in full, a single incorrect support is enough to invalidate a problem, so per-block selection accuracy compounds sharply into plan success. Excluding blocks a method leaves unplaced, the two open-weight models select the correct view at or barely above chance under \formalizer{} and \planner{} ($49.9\%$--$61.4\%$). \gemini{} resolves conflicts far better, at $92.9\%$ and $91.4\%$. \method{} eliminates the gap entirely, reaching $99.8\%$--$100\%$ for all three models: an open-weight model under \method{} fuses views more reliably than \gemini{} does under either existing method, despite the substantial difference in underlying capability.

\paragraph{Contribution of the reliability bonus.} With all observations and other components fixed, removing the reliability bonus lowers plan success on BW-PO-DEPTH from 97\% to 80\% for \gemma{}, 95\% to 79\% for \qwen{}, and 100\% to 83\% for \gemini{}.

\paragraph{Scaling exposes a recall bottleneck in visual state reconstruction.} As BW-PO scales from 6 to 100 blocks, performance degrades because the VLM omits an increasing share of the scene. At 100 blocks, the open-weight models under \method{} retain high average precision but very low recall across the initial and goal states: $92.3\%$ against $9.7\%$ for \gemma{}, and $77.1\%$ against $5.1\%$ for \qwen{} (complete metrics in Appendix Table~\ref{tab:precision-recall}). The relations that are represented are therefore usually correct, but most blocks are absent altogether, and since a valid plan requires both states in full, these omissions drive plan success to zero. The limiting factor in this regime is thus completeness of scene reconstruction rather than local spatial reasoning. \gemini{} sustains $98.9\%$ recall at the same scale, indicating that what limits performance at this scale is the model's visual extraction capacity.

\paragraph{Output token cost.} Measured as generated tokens per successful plan, \method{} is the least expensive method in 11 of the 12 model--dataset cells where the comparison is defined, with a median reduction of 1.8$\times$ over the stronger baseline that widens to 44$\times$ on BW-PO-100. Full results in Appendix Table~\ref{tab:token-cost}.

\paragraph{Prompt-level priors do not close the gap.} Our baselines follow their standard definitions: \planner{} receives the available actions and \formalizer{} the predicates, without further structural guidance. With priors stating that all blocks are the same size, that support is vertical and discrete, and that a block can be supported only by one other block or the table, \planner{} and \formalizer{} reach 0\% and 0\% with \gemma{} and 3\% and 0\% with \qwen{} on BW-PO-DEPTH, against 97\% and 95\% for \method{}.

\begin{table}[t]
\centering
\small
\begin{tabular}{@{}lccc@{}}
\toprule
Dataset & Planner & Formalizer & Perceiver \\
\midrule
BW-PO-DEPTH & 0\% & 0\% & \textbf{13\%} \\
BW-PO-6     & 0\% & 5\% & \textbf{28\%} \\
BW-PO-12    & 0\% & 0\% & \textbf{2\%} \\
BW-PO-20    & \textbf{0\%} & \textbf{0\%} & \textbf{0\%} \\
\bottomrule
\end{tabular}
\caption{Plan success rates with Gemma-4-12B.}
\label{tab:gemma12b}
\end{table}

\paragraph{The trend holds for a weaker model.} We use strong VLMs to show that fusion fails even with high perceptual capacity. With the smaller Gemma-4-12B \citep{gemmateam2026gemma4}, the ordering is preserved (Table~\ref{tab:gemma12b}); we do not evaluate larger scenes, as performance has already reached zero. The reduced gains show that external fusion cannot compensate for weak perception: it helps only once local claims are correct.

\section{Conclusion}
\label{sec:conclusion}
We introduced BW-PO and \method{}. \method{} matches or exceeds \planner{} and \formalizer{} on every model--dataset pair, at lower token cost in 11 of 12 cells where the comparison is defined. Under depth ambiguity, \method{} selects the correct view with near-perfect accuracy, suggesting that the bottleneck in this setting is cross-view fusion rather than local extraction. When fusion is deterministic at 100 blocks, the open-weight models recover only about 10\% of placements while \gemini{} sustains 98.9\% recall.

\method{} does not eliminate the visual-planning bottleneck; it shifts it. Solver-based fusion is near-perfect once the necessary evidence is extracted, but recall under dense scenes remains the key constraint separating frontier models from open-weight models on our hardest datasets.

\section*{Limitations}
\label{sec:limitations}

Our evaluation assumes access to the planning domain and the complete object inventory. This design isolates the capability targeted by BW-PO: recovering initial- and goal-state relations from partially observable visual inputs and using them to produce a valid plan. The benchmark therefore evaluates relational state extraction rather than open-world object discovery, domain induction, or the recovery of missing object identities.

Both BW-PO and \method{} are tied to the vertical-stack structure of Blocksworld: the benchmark induces partial observability through a specific mechanism, one block occluding another along the camera axis, and our fuser encodes matching assumptions of discrete stack indices, direct vertical support, and mutually exclusive cell occupancy. The views are also fixed, so BW-PO does not test active perception, which could resolve ambiguity where cameras can move. Additional views, however, require no change to the method: their claims enter the same tallies and optimization objective. The paradigm itself is not domain-specific, requiring only that local perceptual claims be expressible as weighted constraints, a condition met by any domain with a discrete relational state. We therefore expect it to transfer, though instantiating it elsewhere requires a richer constraint model whose cost we do not measure. Which bottleneck dominates in such domains is a separate question: our fusion result comes from the depth-ambiguity setting, while at high object counts the limiting factor is instead the completeness of the VLM's scene reconstruction.

BW-PO problems are constructed by pairing photographed states, so each state serves as both an initial and a goal state across the dataset. The $110$ problems within each dataset are consequently not statistically independent, and reported success rates should be read as performance over the benchmark instances rather than as estimates derived from $110$ independent scene samples.

Scaling introduces a further confound. Increasing the number of blocks reduces the pixels available per object and increases the difficulty of reading their numerical identifiers. Degradation at larger scales may therefore reflect not only the greater number of objects and relations to be recovered, but also reduced visual resolution and increased recognition difficulty. Future variants could control object size and image resolution independently of scene complexity.

\bibliography{custom}

@article{hu2023look,
      title={Look Before You Leap: Unveiling the Power of {GPT-4V} in Robotic Vision-Language Planning},
      author={Yingdong Hu and Fanqi Lin and Tong Zhang and Li Yi and Yang Gao},
      journal={arXiv preprint arXiv:2311.17842},
      year={2023}
    }

@INPROCEEDINGS{10611112,
  author={Shirai, Keisuke and Beltran-Hernandez, Cristian C. and Hamaya, Masashi and Hashimoto, Atsushi and Tanaka, Shohei and Kawaharazuka, Kento and Tanaka, Kazutoshi and Ushiku, Yoshitaka and Mori, Shinsuke},
  booktitle={2024 IEEE International Conference on Robotics and Automation (ICRA)}, 
  title={Vision-Language Interpreter for Robot Task Planning}, 
  year={2024},
  volume={},
  number={},
  pages={2051-2058},
  doi={10.1109/ICRA57147.2024.10611112}}

@inproceedings{merler-etal-2026-viplan,
    title = "{V}i{P}lan: A Benchmark for Visual Planning with Symbolic Predicates and Vision-Language Models",
    author = "Merler, Matteo  and
      Dainese, Nicola  and
      Alakuijala, Minttu  and
      Bonetta, Giovanni  and
      Ferrazzi, Pietro  and
      Tian, Yu  and
      Magnini, Bernardo  and
      Marttinen, Pekka",
    editor = "Lee, Hung-Yi  and
      Dev, Sunipa  and
      Li, Jessy  and
      Elliott, Desmond",
    booktitle = "Findings of the {A}ssociation for {C}omputational {L}inguistics: {EMNLP} 2026",
    month = oct,
    year = "2026",
    address = "Budapest, Hungary",
    publisher = "Association for Computational Linguistics",
    url = "https://aclanthology.org/2026.findings-emnlp.1017/",
    doi = "10.18653/2026.findings-emnlp.1017",
    pages = "21469--21516",
    ISBN = "979-8-89176-462-0"
}

@misc{he2025visionlanguagemodelsplan,
      title={Vision Language Models Cannot Plan, but Can They Formalize?}, 
      author={Muyu He and Yuxi Zheng and Yuchen Liu and Zijian An and Bill Cai and Jiani Huang and Lifeng Zhou and Feng Liu and Ziyang Li and Li Zhang},
      year={2025},
      eprint={2509.21576},
      archivePrefix={arXiv},
      primaryClass={cs.CL},
      url={https://arxiv.org/abs/2509.21576}, 
}

@inproceedings{
    nayak2024longhorizon,
    title={Long-Horizon Planning for Multi-Agent Robots in Partially Observable Environments},
    author={Siddharth Nayak and Adelmo Morrison Orozco and Marina Ten Have and Jackson Zhang and Vittal Thirumalai and Darren Chen and Aditya Kapoor and Eric Robinson and Karthik Gopalakrishnan and James Harrison and Anuj Mahajan and brian ichter and Hamsa Balakrishnan},
    booktitle={The Thirty-eighth Annual Conference on Neural Information Processing Systems},
    year={2024},
    url={https://openreview.net/forum?id=Y1rOWS2Z4i}
}

@misc{yun2026integratingmultimodallargelanguage,
      title={Integrating Multimodal Large Language Model Knowledge into Amodal Completion}, 
      author={Heecheol Yun and Eunho Yang},
      year={2026},
      eprint={2603.28333},
      archivePrefix={arXiv},
      primaryClass={cs.CV},
      url={https://arxiv.org/abs/2603.28333}, 
}

@INPROCEEDINGS{pothiraj2025capture,
  author={Pothiraj, Atin and Stengel-Eskin, Elias and Cho, Jaemin and Bansal, Mohit},
  booktitle={2025 IEEE/CVF International Conference on Computer Vision (ICCV)}, 
  title={{CAPTURe}: Evaluating Spatial Reasoning in Vision Language Models via Occluded Object Counting}, 
  year={2025},
  volume={},
  number={},
  pages={8001-8010},
  doi={10.1109/ICCV51701.2025.00750}
}

@misc{vo2026depthbettertoldshown,
      title={When Depth Is Better Told Than Shown: Depth-Ordinal Prompting for Vision-Language Spatial Reasoning}, 
      author={Quynh Vo and Phuc Dao and Cong-Duy Nguyen and Thong Nguyen},
      year={2026},
      eprint={2607.11173},
      archivePrefix={arXiv},
      primaryClass={cs.CV},
      url={https://arxiv.org/abs/2607.11173}, 
}

@misc{sengupta2026visionlanguagemodelscountsynthetic,
      title={Can Vision-Language Models Count? A Synthetic Benchmark and Analysis of Attention-Based Interventions}, 
      author={Saurav Sengupta and Nazanin Moradinasab and Jiebei Liu and Donald E. Brown},
      year={2026},
      eprint={2511.17722},
      archivePrefix={arXiv},
      primaryClass={cs.CV},
      url={https://arxiv.org/abs/2511.17722}, 
}

@inproceedings{
wang2025muirbench,
title={{MuirBench}: A Comprehensive Benchmark for Robust Multi-image Understanding},
author={Fei Wang and Xingyu Fu and James Y. Huang and Zekun Li and Qin Liu and Xiaogeng Liu and Mingyu Derek Ma and Nan Xu and Wenxuan Zhou and Kai Zhang and Tianyi Lorena Yan and Wenjie Jacky Mo and Hsiang-Hui Liu and Pan Lu and Chunyuan Li and Chaowei Xiao and Kai-Wei Chang and Dan Roth and Sheng Zhang and Hoifung Poon and Muhao Chen},
booktitle={The Thirteenth International Conference on Learning Representations},
year={2025},
url={https://openreview.net/forum?id=TrVYEZtSQH}
}

@inproceedings{
feng2026seeing,
title={Seeing Across Views: Benchmarking Spatial Reasoning of Vision-Language Models in Robotic Scenes},
author={ZhiYuan Feng and Zhaolu Kang and Qijie Wang and Zhiying Du and Jiongrui Yan and Shi Shubin and Chengbo Yuan and Huizhi Liang and Yu Deng and Qixiu Li and Rushuai Yang and Ruichuan An and Leqi Zheng and Weijie Wang and Shuang Chen and Sicheng Xu and Yaobo Liang and Jiaolong Yang and Baining Guo},
booktitle={The Fourteenth International Conference on Learning Representations},
year={2026},
url={https://openreview.net/forum?id=jXDZJAfRZB}
}

@misc{ipc1998,
    title = {International Planning Competition},
    url = {https://www.icaps-conference.org/competitions},
    author = {IPC},
    year = {1998}
}

@article{Helmert_2006,
   title={The Fast Downward Planning System},
   volume={26},
   ISSN={1076-9757},
   url={http://dx.doi.org/10.1613/jair.1705},
   DOI={10.1613/jair.1705},
   journal={Journal of Artificial Intelligence Research},
   publisher={AI Access Foundation},
   author={Helmert, M.},
   year={2006},
   month=July, pages={191–246}
}

@InProceedings{10.1007/978-3-540-78800-3_24,
    author="de Moura, Leonardo
    and Bj{\o}rner, Nikolaj",
    editor="Ramakrishnan, C. R.
    and Rehof, Jakob",
    title="Z3: An Efficient {SMT} Solver",
    booktitle="Tools and Algorithms for the Construction and Analysis of Systems",
    year="2008",
    publisher="Springer Berlin Heidelberg",
    address="Berlin, Heidelberg",
    pages="337--340",
    isbn="978-3-540-78800-3"
}

@InProceedings{10.1007/978-3-662-46681-0_14,
    author="Bj{\o}rner, Nikolaj
    and Phan, Anh-Dung
    and Fleckenstein, Lars",
    editor="Baier, Christel
    and Tinelli, Cesare",
    title="$\nu$Z - An Optimizing {SMT} Solver",
    booktitle="Tools and Algorithms for the Construction and Analysis of Systems",
    year="2015",
    publisher="Springer Berlin Heidelberg",
    address="Berlin, Heidelberg",
    pages="194--199",
    isbn="978-3-662-46681-0"
}

@article{unified_planning_softwarex2025,
  title = {Unified Planning: Modeling, manipulating and solving AI planning problems in Python},
  author = {Andrea Micheli and Arthur Bit-Monnot and Gabriele R{\"o}ger and Enrico Scala and Alessandro Valentini and Luca Framba and Alberto Rovetta and Alessandro Trapasso and Luigi Bonassi and Alfonso Emilio Gerevini and Luca Iocchi and Felix Ingrand and Uwe Köckemann and Fabio Patrizi and Alessandro Saetti and Ivan Serina and Sebastian Stock},
  journal = {SoftwareX},
  volume = {29},
  pages = {102012},
  year = {2025},
  issn = {2352-7110},
  doi = {https://doi.org/10.1016/j.softx.2024.102012},
  url = {https://www.sciencedirect.com/science/article/pii/S2352711024003820}
}

@misc{gemmateam2026gemma4,
      title={Gemma 4 Technical Report}, 
      author={{Gemma Team}},
      year={2026},
      eprint={2607.02770},
      archivePrefix={arXiv},
      primaryClass={cs.CL},
      url={https://arxiv.org/abs/2607.02770}, 
}

@misc{qwen3.5,
    title  = {{Qwen3.5}: Towards Native Multimodal Agents},
    author = {{Qwen Team}},
    month  = {February},
    year   = {2026},
    url    = {https://qwen.ai/blog?id=qwen3.5}
}

@inproceedings{kwon2023efficient,
  title={Efficient Memory Management for Large Language Model Serving with {PagedAttention}},
  author={Woosuk Kwon and Zhuohan Li and Siyuan Zhuang and Ying Sheng and Lianmin Zheng and Cody Hao Yu and Joseph E. Gonzalez and Hao Zhang and Ion Stoica},
  booktitle={Proceedings of the ACM SIGOPS 29th Symposium on Operating Systems Principles},
  year={2023}
}

@inproceedings{pan-etal-2024-langnav,
    title = "{L}ang{N}av: Language as a Perceptual Representation for Navigation",
    author = "Pan, Bowen  and
      Panda, Rameswar  and
      Jin, SouYoung  and
      Feris, Rogerio  and
      Oliva, Aude  and
      Isola, Phillip  and
      Kim, Yoon",
    editor = "Duh, Kevin  and
      Gomez, Helena  and
      Bethard, Steven",
    booktitle = "Findings of the {A}ssociation for {C}omputational {L}inguistics: {NAACL} 2024",
    month = jun,
    year = "2024",
    address = "Mexico City, Mexico",
    publisher = "Association for Computational Linguistics",
    url = "https://aclanthology.org/2024.findings-naacl.60/",
    doi = "10.18653/v1/2024.findings-naacl.60",
    pages = "950--974"
}

@misc{zhou2023navgptexplicitreasoningvisionandlanguage,
      title={{NavGPT}: Explicit Reasoning in Vision-and-Language Navigation with Large Language Models}, 
      author={Gengze Zhou and Yicong Hong and Qi Wu},
      year={2023},
      eprint={2305.16986},
      archivePrefix={arXiv},
      primaryClass={cs.CV},
      url={https://arxiv.org/abs/2305.16986}, 
}

\appendix

\section{Dataset Construction}
\label{app:dataset_construction}

\paragraph{Problem generation.} For each dataset we capture $11$ distinct physical states, each photographed from a front and a side viewpoint, and form problems from all ordered pairs of distinct states. This yields $P(11,2)=11\times10=110$ planning problems per dataset. Ordering matters, since $(s_i,s_j)$ and $(s_j,s_i)$ pose different problems, and every state serves as both an initial and a goal state across the dataset.

\paragraph{Difficulty filtering.} In the six-block datasets the reachable configuration space is small enough that randomly paired states frequently admit trivial or near-trivial solutions. For these datasets we therefore capture additional states, discard every pair whose optimal ground-truth plan contains fewer than five actions, and sample $110$ problems from those remaining, preserving a consistent dataset size across all of BW-PO.

\paragraph{Block identity.} We retain the same six block colors at every value of $N$, so colors are necessarily reused as the number of blocks grows. Each physical block is therefore marked with a unique numerical identifier, and blocks are referred to by the color--identifier pair throughout. Color alone is sufficient to name a block only in the six-block datasets.

\section{Gripper Resolution}
\label{app:gripper}

The gripper state is resolved independently of the block placement. Each view contributes at most one vote
\[
    g_v\in O\cup\{\emptyset,\bot\},
\]
where $\emptyset$ denotes an empty gripper and $\bot$ denotes abstention. Within a view, a report naming a held block takes precedence over a conflicting empty-hand statement, and a view that states only that the gripper is full, without identifying the held block, abstains. A named block is accepted when the other view agrees or abstains; conflicting named votes and all remaining cases default to an empty gripper:
\begin{equation*}
    g=
    \begin{cases}
    x, & g_{\mathrm{front}}=g_{\mathrm{side}}=x\in O,\\
    x, & g_v=x\in O\ \wedge\ g_{v'}=\bot,\\
    \emptyset, & \text{otherwise.}
\end{cases}
\end{equation*}

A held block is detached from the reconstructed arrangement and receives no spatial predicates of its own. If another block was reported as resting on the held block, that block is instead placed on the table; any block reported beneath the held block is treated as clear. The rule is defined for generality: it is required in embodied settings where the gripper is observable, and on our data it acts only to reject unsupported \texttt{holding} claims.

\section{Inference Details}
\label{app:inference}

The open-weight models are served locally with vLLM \citep{kwon2023efficient} on eight NVIDIA H100 GPUs, using the instruction-tuned checkpoints \texttt{google/\allowbreak gemma-4-12b-it}, \texttt{google/\allowbreak gemma-4-31b-it}, and \texttt{Qwen/\allowbreak Qwen3.5-122B-A10B}. \gemini{} refers to \texttt{gemini-3.1-pro-preview}, accessed through the Gemini Enterprise Agent Platform. We use the default inference hyperparameters for each model, except that temperature is set to zero to improve reproducibility and the reasoning budget is set to 32,768 tokens. This budget was selected through a preliminary sweep with additional headroom; responses truncated before producing complete, parseable output are scored as incorrect. Images are resized to $256\times256$ with no other preprocessing.

\begin{table*}[t]
\centering
\small
\setlength{\tabcolsep}{5pt}
\resizebox{\textwidth}{!}{%
\begin{tabular}{@{}l*{3}{ccc}@{}}
\toprule
\multicolumn{1}{c}{}
& \multicolumn{3}{c}{Gemma-4-31B}
& \multicolumn{3}{c}{Qwen3.5-122B-A10B}
& \multicolumn{3}{c}{Gemini-3.1-Pro} \\
\cmidrule(lr){2-4}
\cmidrule(lr){5-7}
\cmidrule(lr){8-10}
\multicolumn{1}{c}{Dataset}
& Planner & Formalizer & Perceiver
& Planner & Formalizer & Perceiver
& Planner & Formalizer & Perceiver \\
\midrule
BW-PO-DEPTH
& 1290 & -- & \textbf{7.9}
& -- & -- & \textbf{9.6}
& 23.0 & 40.1 & \textbf{21.1} \\
BW-PO-6
& 9.7 & 10.1 & \textbf{2.7}
& 6.9 & 7.2 & \textbf{3.5}
& 3.9 & \textbf{3.6} & 4.0 \\
BW-PO-12
& 20.5 & 67.0 & \textbf{5.4}
& 48.0 & 47.3 & \textbf{32.0}
& 18.0 & 10.9 & \textbf{8.4} \\
BW-PO-20
& 932.9 & 1244.1 & \textbf{70.3}
& -- & -- & \textbf{613.9}
& 80.8 & 18.5 & \textbf{11.7} \\
BW-PO-40
& -- & -- & \textbf{3943.5}
& -- & -- & --
& 91.5 & 25.1 & \textbf{15.0} \\
BW-PO-100
& -- & -- & --
& -- & -- & --
& 3564.4 & 2068.9 & \textbf{47.5} \\
\bottomrule
\end{tabular}%
}
\caption{Mean generated tokens per successful plan, in thousands. Dashes indicate that no plan was validated successfully, leaving the ratio undefined.}
\label{tab:token-cost}
\end{table*}

\section{Error Analysis on BW-PO-DEPTH}
\label{app:error_analysis}

We reduce each method's output to a \emph{support map}: for every block, it reports the single object it rests on, whether another block, the table, or the gripper. This is the only claim all three methods make in common, and it is scored against ground truth over $110$ problems $\times$ 2 states $\times$ $6$ blocks $=1320$ placements. \formalizer{} and \method{} are read directly from their emitted PDDL. \planner{} emits no state at all, so its support map must be recovered from its plan. Each action commits the plan to a belief about the block in the initial state: \texttt{(unstack $x$ $y$)} entails that $x$ was on $y$, and \texttt{(pick-up $x$)} that $x$ was on the table. Taking the first action to touch each block and simulating forward from there yields the weakest initial state under which the plan is executable. Blocks the plan never touches carry no such commitment and are excluded from its score. We report \emph{coverage}, the share of the $1320$ placements that remain, which is $39.5\%$ for \gemma{}, $36.2\%$ for \qwen{}, and $63.6\%$ for \gemini{}. Because the excluded blocks are precisely those the plan avoided, \planner{}'s accuracy is measured on a self-selected subset and should be read as an upper bound.

Ambiguity labels come from a single \method{} run with \gemini{} that records each view's description before fusion. A placement is \emph{ambiguous} when both views place the block and disagree, and \emph{unambiguous} otherwise. Re-deriving the labels from a \gemma{} run agrees with the \gemini{}-derived labels on $97.9\%$ of the $1320$ placements, so the split reflects the images rather than the model that produced it. Because a block that both views describe identically but wrongly is labelled unambiguous, the ambiguous set is a lower bound. Under these labels the $1320$ placements divide into $389$ that both views place identically, $519$ that exactly one view places, and $412$ ambiguous ones; none are placed by neither view. Every uncontradicted claim is correct, and in all $412$ ambiguous cases the true support is one of the two competing claims, so each disagreement is a binary choice with the answer present in the evidence. Both properties follow from how BW-PO-DEPTH is constructed: each view is ambiguous on its own, yet together they uniquely determine the state, so under correct perception the problem would otherwise be unsolvable from the evidence. The reliability bonus of Section~\ref{sec:methods} relies on this structure. Selection on an ambiguous placement is therefore one of four outcomes, in which a method names the support one view claims, the support the other claims, a third support neither claims, or declines to place the block; the rates reported in Section~\ref{sec:experiments} exclude the last case.

\begin{table*}[t]
\centering
\small
\setlength{\tabcolsep}{7pt}
\renewcommand{\arraystretch}{1.15}
\begin{tabular}{@{}ll rr rr rr@{}}
\toprule
& & \multicolumn{2}{c}{\gemma{}}
  & \multicolumn{2}{c}{\qwen{}}
  & \multicolumn{2}{c}{\gemini{}} \\
\cmidrule(lr){3-4} \cmidrule(lr){5-6} \cmidrule(lr){7-8}
Dataset & Method & Precision & Recall & Precision & Recall & Precision & Recall \\
\midrule
\multirow{3}{*}{BW-PO-6}
  & Planner    & 84.4 & 62.3 & \textbf{100.0} & 75.5 & 95.5 & 71.1 \\
  & Formalizer & 79.5 & 79.4 & \textbf{100.0} & \textbf{100.0} & \textbf{100.0} & \textbf{100.0} \\
  & Perceiver  & \textbf{100.0} & \textbf{99.2} & \textbf{100.0} & 99.6 & \textbf{100.0} & \textbf{100.0} \\
\addlinespace
\multirow{3}{*}{BW-PO-12}
  & Planner    & 81.3 & 70.6 & 82.3 & 72.0 & 97.6 & 84.6 \\
  & Formalizer & 87.3 & 75.9 & 93.0 & \textbf{92.2} & 99.5 & 99.5 \\
  & Perceiver  & \textbf{99.7} & \textbf{98.3} & \textbf{98.0} & 84.0 & \textbf{100.0} & \textbf{100.0} \\
\addlinespace
\multirow{3}{*}{BW-PO-20}
  & Planner    & 41.3 & 38.3 & 45.6 & 39.0 & 89.3 & 83.4 \\
  & Formalizer & 83.5 & 65.0 & 81.0 & \textbf{65.7} & 99.8 & 99.8 \\
  & Perceiver  & \textbf{94.0} & \textbf{86.7} & \textbf{93.8} & 56.3 & \textbf{100.0} & \textbf{100.0} \\
\addlinespace
\multirow{3}{*}{BW-PO-40}
  & Planner    & 35.6 & 30.9 & 9.0 & 6.6 & 77.6 & 76.2 \\
  & Formalizer & 70.1 & 46.1 & 52.5 & \textbf{36.7} & 95.0 & 95.0 \\
  & Perceiver  & \textbf{92.6} & \textbf{52.0} & \textbf{76.4} & 35.1 & \textbf{99.7} & \textbf{99.7} \\
\addlinespace
\multirow{3}{*}{BW-PO-100}
  & Planner    & 23.8 & 8.3 & 6.3 & 2.0 & 8.7 & 4.1 \\
  & Formalizer & 34.7 & \textbf{19.3} & 24.5 & \textbf{12.0} & 59.9 & 40.7 \\
  & Perceiver  & \textbf{92.3} & 9.7 & \textbf{77.1} & 5.1 & \textbf{99.8} & \textbf{98.9} \\
\bottomrule
\end{tabular}
\caption{Planner follows the same recovery as described in Appendix~\ref{app:error_analysis} in order to compute precision and recall. Each value is the mean of the initial state and goal state score, reported in percent.}
\label{tab:precision-recall}
\end{table*}

\begin{table*}[t]
\centering
\small
\setlength{\tabcolsep}{5pt}
\resizebox{\textwidth}{!}{%
\begin{tabular}{@{}l*{4}{cc}@{}}
\toprule
\multicolumn{1}{c}{}
& \multicolumn{4}{c}{Gemma-4-31B}
& \multicolumn{4}{c}{Qwen3.5-122B-A10B} \\
\cmidrule(lr){2-5}
\cmidrule(lr){6-9}
& \multicolumn{2}{c}{Plan success rate}
& \multicolumn{2}{c}{Mean tokens per problem }
& \multicolumn{2}{c}{Plan success rate}
& \multicolumn{2}{c}{Mean tokens per problem} \\
\cmidrule(lr){2-3}\cmidrule(lr){4-5}\cmidrule(lr){6-7}\cmidrule(lr){8-9}
\multicolumn{1}{c}{Dataset}
& Reasoning & No reasoning & Reasoning & No reasoning
& Reasoning & No reasoning & Reasoning & No reasoning \\
\midrule
BW-PO-DEPTH
& 97\% & \textbf{100\%} & 7.7 & 0.4
& \textbf{95\%} & 82\% & 9.0 & 1.8 \\
BW-PO-6
& \textbf{99\%} & 96\% & 2.7 & 0.3
& \textbf{100\%} & 44\% & 3.5 & 1.3 \\
BW-PO-12
& \textbf{86\%} & 38\% & 4.7 & 0.5
& \textbf{51\%} & 7\% & 16.3 & 4.8 \\
BW-PO-20
& \textbf{22\%} & 15\% & 15.3 & 1.0
& \textbf{6\%} & 1\% & 39.1 & 13.3 \\
BW-PO-40
& \textbf{1\%} & 0\% & 35.8 & 16.8
& 0\% & 0\% & 49.4 & 24.9 \\
BW-PO-100
& 0\% & 0\% & 58.0 & 12.2
& 0\% & 0\% & 58.9 & 34.3 \\
\bottomrule
\end{tabular}%
}
\caption{\method{} with each open-weight model's reasoning mode enabled and disabled. Unlike Table~\ref{tab:token-cost}, token counts here are averaged over all problems rather than over successful plans, since the no-reasoning configurations succeed too rarely for the per-success ratio to be meaningful.}
\label{tab:reasoning_mode}
\end{table*}

\section{Effect of Reasoning Mode}
\label{app:reasoning_mode}

The restricted perception task used by \method{} asks the VLM only to report local spatial observations within a single view, which raises the question of whether the open-weight models require their reasoning modes in this setting. We therefore re-evaluate \gemma{} and \qwen{} under \method{} with reasoning disabled, holding all other inference settings fixed.

The effect depends on scene size rather than on perceptual difficulty. On BW-PO-DEPTH, \gemma{} is marginally \emph{better} without reasoning, reaching 100\% against 97\% while generating roughly 20$\times$ fewer tokens; \qwen{} loses 13 points on the same dataset. As object count grows, reasoning becomes necessary: \gemma{} falls from 86\% to 38\% on BW-PO-12, and \qwen{} from 100\% to 44\% on BW-PO-6 and from 51\% to 7\% on BW-PO-12. We therefore find no setting in which reasoning can be disabled uniformly, and report all main-text results with each model's default reasoning mode enabled.

\end{document}